\def\year{2022}\relax
\documentclass[letterpaper]{article} %
\usepackage{aaai22}  %

\usepackage{times}  %
\usepackage{helvet}  %
\usepackage{courier}  %
\usepackage[hyphens]{url}  %
\usepackage{graphicx} %
\usepackage{natbib}  %
\usepackage{caption} %
\DeclareCaptionStyle{ruled}{labelfont=normalfont,labelsep=colon,strut=off} %
\usepackage[title]{appendix}

\usepackage{color}

\usepackage{algorithm}
\usepackage[noend]{algpseudocode}
\usepackage{booktabs} %
\usepackage[colorlinks=true,linkcolor=blue]{hyperref}%
\hypersetup{
  citecolor=gray}
\usepackage{bbm}
\usepackage{amsmath,amsthm,amsfonts,amssymb,amscd}
\usepackage{csvsimple}
\usepackage{xcolor}
\definecolor{blue}{RGB}{61, 133, 198}
\usepackage[normalem]{ulem} %
\usepackage{array}
\usepackage{csquotes} %

\usepackage{newfloat}
\usepackage{listings}
\floatstyle{ruled}
\newfloat{listing}{tb}{lst}{}
\floatname{listing}{Listing}
\title{One Map Does Not Fit All: \\
Evaluating Saliency Map Explanation on Multi-Modal Medical Images}
\title{Evaluating Explainable AI on Medical Image Task: \\a Clinical Requirement Oriented Approach}
\title{Evaluating Explainable AI on a Multi-Modal Medical Imaging Task: \\Can Existing Algorithms Fulfill Clinical Requirements?}
\author{
}

\author {
    Weina Jin,\textsuperscript{\rm 1}
    Xiaoxiao Li, \textsuperscript{\rm 2}
    Ghassan Hamarneh \textsuperscript{\rm 1}
}
\affiliations {
    \textsuperscript{\rm 1} School of Computing Science, Simon Fraser University\\
    \textsuperscript{\rm 2} Department of Electrical and Computer Engineering, The University of British Columbia\\
    weinaj@sfu.ca, xiaoxiao.li@ece.ubc.ca, hamarneh@sfu.ca
}

\usepackage{bibentry}

\begin{document}

\maketitle

\begin{abstract}
Being able to explain the prediction to clinical end-users is a necessity to leverage the power of artificial intelligence (AI) models for clinical decision support. For medical images, a feature attribution map, or heatmap, is the most common form of explanation that highlights important features for AI models' prediction. However, it is unknown how well heatmaps perform on explaining decisions on multi-modal medical images, where each image modality or channel visualizes distinct clinical information of the same underlying biomedical phenomenon.
Understanding such modality-dependent features is essential for clinical users' interpretation of AI decisions.
To tackle this clinically important but technically ignored problem, we propose the modality-specific feature importance (MSFI) metric. 
It encodes clinical image and explanation interpretation patterns of modality prioritization and modality-specific feature localization.
We conduct a clinical requirement-grounded, systematic evaluation 
using computational methods and 
a clinician user study. Results show that the examined 16 heatmap algorithms failed to fulfill clinical requirements to correctly indicate AI model decision process or decision quality.
The evaluation and MSFI metric can guide the design and selection of XAI algorithms to meet clinical requirements on multi-modal explanation.
\end{abstract}

\section{Introduction}\label{intro}
Being able to explain decisions to users is a sought-after quality of artificial intelligence (AI) or deep learning (DL) based predictive models, particularly when deploying them in high-stakes real-world applications, such as clinical decision support systems~\cite{Jin_2020,He2019}.
Explanations can help clinical end-users verify models' decision~\cite{10.1145/2939672.2939778}, resolve disagreements with AI during decision discrepancy~\cite{10.1145/3359206}, 
calibrate their trust in AI assistance~\cite{7349687}, 
 and ultimately facilitate doctor-AI communication and collaboration to leverage the strengths of both~\cite{Topol2019}.

To understand AI decision on medical imaging tasks,  the most common and clinical end-user-friendly explanation is a heatmap or feature attribution map~\cite{Reyes2020}. It highlights the important regions on the input image for the model's prediction. Despite many explanation algorithms have been proposed in the explainable AI (XAI) and computer vision communities~\cite{simonyan2014deep,NIPS2017_8a20a862,8237336,10.1145/2939672.2939778}, there is a lack of systematic evaluation on their correctness and usefulness in medical imaging tasks. It is an open question to evaluate if the existing XAI algorithms can fulfill clinical requirements, given these methods were originally proposed on natural images.

However, XAI evaluation is notoriously challenging and still immature, due to complex human factors and application scenarios.
Most of the existing evaluation desiderata and metrics are chosen or proposed by AI practitioners, with little involvement of model end-users in this process~\cite{jin2021euca}. Such an engineer-centered evaluation paradigm may be problematic in domains that require experts' knowledge to define the problem, such as medicine. To tackle this issue, we conduct a clinical requirement-grounded, systematic evaluation on a real and common clinical task with multi-modal medical images. 
With close collaboration with physicians, we first formulate the \textbf{\textit{clinically-important-but-technically-ignored problem}} of explaining on multi-modal medical images, then define evaluation desiderata and metrics based on \textbf{\textit{clinical requirements}}.

To address the XAI evaluation problem in medical imaging tasks, this work focuses on its general form of \textit{multi-modal medical images}, which are widely used in clinical settings for critical decision-support, such as multi-pulse sequence magnetic resonance imaging (MRI), PET-CT, and multi-stained pathological images. Interpreting information from multi-modal data is a complex process in clinical practice. Doctors usually compare and combine modality-specific information to reason about diagnosis and differential diagnosis. For instance, in a radiology report on MRI, radiologists usually observe and describe \textit{anatomical} structures in T1 modality, and \textit{pathological} changes in T2 modality~\cite{cochard_netter_2012, Bitar2006}; doctors can infer the composition of a lesion (such as fat, hemorrhage, protein, fluid) by combining its signals from different MRI modalities~\cite{Patel2016}. In addition, some imaging modalities are particularly crucial for the diagnosis and management of certain diseases~\cite{Lansberg2000}.%

Existing XAI methods~\cite{simonyan2014deep,NIPS2017_8a20a862,8237336,10.1145/2939672.2939778} are typically not designed for clinical purposes. For example, the current XAI in medical imaging analysis (MIA) mainly focuses on explaining models on single image modality, which conforms with natural image explanation settings, but over-simplifies or ignores the above complex clinical decision process with medical images. Further, those XAI methods may not consider end-users' image and explanation interpretation patterns by design.
Two clinical patterns on multi-modal explanation were extracted based on our user study with physicians (\S\ref{user_study}): the heatmap needs to \textbf{1}) highlight important modalities and to \textbf{2}) localize important features for model prediction.

In this work, we formulate a novel problem of multi-modal explanation to the technical community, and present a systematic evaluation on this problem.
The evaluation inspects two main clinical requirements of explanation: how \textit{faithful} the explanation describes the AI model's internal decision process, and how the human assessment of the explanation \textit{plausibility} is indicative for the model's decision quality.
We propose a computational metric \underline{m}odality-\underline{s}pecific \underline{f}eature \underline{i}mportance (MSFI) 
that summarizes the above clinical patterns on multi-modal explanation.
We then conduct a systematic evaluation on 16 XAI methods that cover the most common activation-, gradient-, and perturbation-based approaches in a brain tumor classification task using multi-modal MRI. Our key contributions are:

\begin{enumerate}
    \item We conduct a systematic evaluation on a medical imaging task, that covers both quantitative and qualitative physician evaluation, and clinical requirements grounded computational evaluation on explanation \textit{faithfulness} and \textit{plausibility}. Results show that existing methods are not proposed to fulfill the clinical requirements of modality-specific medical imaging explanation.
    \item We formulate and tackle the novel and clinically significant problem of multi-modal image explanation, as it is the generalized form of single-modal image explanation.
    \item We propose the computational evaluation metric MSFI, which automates the human assessment process by incorporating the clinical patterns of modality prioritization and feature localization. 
\end{enumerate}

\section{Related Work}
\subsection{Clinical Requirement-Grounded  \quad\quad\quad \quad XAI Evaluation Desiderata and Metrics}
Existing surveys~\cite{Sokol2020,10.1145/3387166,VILONE202189,8400040} outline many desiderata as proxies for real-world outcomes to guide the design and evaluation of XAI algorithms, such as correctness, robustness~\cite{DBLP:journals/corr/abs-1806-08049}, simulatability~\cite{hase-bansal-2020-evaluating}.
We develop the critical task-specific evaluation desiderata, which are grounded in the literature on explanation correctness~\cite{jacovi-goldberg-2020-towards}, and in the understandings of the clinical utility of explanation via prior~\cite{jin2021euca} and our clinical user study: the primary objectives of explanation in critical tasks are to enable users to \textbf{1}) understand why, how, and when AI works and does not work. As a prerequisite for users to build a precise mental model of AI, the explanation should \textit{faithfully} reflect the model decision process. \textbf{2}) Explanation enables users to verify AI decisions and identify potential errors via human judgment on how \textit{plausible} the explanation is. We summarize computational metrics in prior work related to the two clinical requirements of explanation:

\textbf{\textit{Faithfulness}} measures how accurately the explanation reflects the model's true decision process. It cannot be measured by human judgment or annotated ground truth encoding human prior knowledge, as humans have no idea about a model's internal decision process. Common evaluation methods include gradually erasing or adding features to input and measuring its effect on model performance~\cite{DBLP:journals/corr/abs-2104-08782,NEURIPS2019_a7471fdc,DBLP:conf/nips/HookerEKK19,7552539,Lundberg2020}, and constructing synthetic datasets with known ground truth features~\cite{pmlr-v80-kim18d}. 

\textbf{\textit{Plausibility}} is the users' assessment of how agreeable the explanation is with their prior knowledge of the task.
It requires human annotated ground truth to reflect human prior knowledge on a given task, such as feature segmentation masks or bounding boxes. Agreement metrics comparing a heatmap with the ground truth mask, such as intersection over union (IoU), are widely used
~\cite{10.1007/978-3-030-32226-7_82,netdissect2017}.

\subsection{XAI Evaluation in Medical Image Analysis}
Although many XAI algorithms have been proposed for or applied in various MIA tasks~\cite{jimaging6060052}, extensive evaluations on their correctness and clinical utility are under-explored. 
In our ongoing review, among 102 works (in Supplementary Material) that apply or propose XAI algorithms for medical imaging tasks, 35\% evaluated the explanation with computational metrics
only; 8\% evaluated via physician user study to verify \textit{explanation plausibility} either quantitatively or qualitatively. Only 5\% have both computational and physician evaluation. 

There are very limited emerging works in which XAI evaluation is the main focus. Recently, \citet{10.1007/978-3-030-63419-3_3} evaluated 13 XAI algorithms on classifying eye diseases using retina images and asked 14 clinicians to rate the heatmaps regarding their clinical relevance (\textit{plausibility}). %
Concurrent with our work, \citet{DESOUZA2021104578} evaluated five gradient-based XAI algorithms 
in classifying early cancer from endoscopic images.
They used computational metrics to measure the agreement between heatmaps and the ground-truth annotations of localized lesion (\textit{plausibility}).
Gradient method outperformed the rest four algorithms that best matches with doctors' ground-truth annotations.

The above-mentioned works either evaluated XAI algorithms in a case-by-case manner, or only addressed computational metrics or doctors' ratings, without utilizing both. To the best of our knowledge, few works conducted both user studies and computational metrics evaluation on XAI for MIA in a systematic manner, nor did they incorporate clinical requirements in the evaluation.
Furthermore, evaluation on the problem of multi-modal medical image explanations is underexplored.
Our work is the first to address these research gaps. 

\section{Clinical Task, Data, and Model}
We present the clinical task, medical dataset, and convolutional neural network (CNN) models prepared for the evaluation.
\paragraph{Clinical Task and Data} As a type of primary brain tumors, gliomas are one of the most devastating cancers. Grading gliomas based on MRI could provide physicians indispensable information on patients' treatment plan and prognosis. AI-based clinical decision support equipped with explanations has the potential to assist neurosurgeons to predict glioma grade and their genetic biomarker status based on brain imaging~\cite{Jin_2020}. 

In our evaluation, we focus on the glioma grading task to classify gliomas into lower-grade (LGG) or high-grade gliomas (HGG). We used the publicly available BraTS 2020 dataset\footnote{Multimodal Brain Tumor Segmentation Challenge \href{http://www.med.upenn.edu/cbica/brats2020/data.html}{www.med.upenn.edu/cbica/brats2020/data.html}} and a BraTS-based synthetic dataset (described in \S\ref{syn}). Both are multi-modal MRI 
with four modalities of T1, T1C (contrast enhancement), T2, and FLAIR.

We chose this task because we have access to clinical collaborators who can provide clinical insights and assessment. In addition, built upon the publicly available BraTS and its associated TCIA dataset which contains rich clinical and genomic labels, the basic tumor grading task can easily be extended to other clinically relevant tasks such as predicting patients' genetic biomarker mutant status or prognosis. 

\subsubsection{Multi-Modality Learning}
Approaches for building CNN models that fuse multi-modal medical images can be divided into three categories: methods that fuse multi-modal features at the \textit{input}-level, \textit{feature}-level, or \textit{decision}-level~\cite{10.1007/978-3-030-32962-4_18}. We focus on the most common setting of multi-modal medical imaging learning tasks: \textit{input}-level multi-modal image fusion~\cite{Shen2017}, in which the multi-modal images are stacked as image channels and fed as input to a deep convolutional neural network. The modality-specific information is fused by summing up the weighted modality value in the first convolutional layer.

Specifically\footnote{Code: \href{http://github.com/weinajin/multimodal_explanation}{github.com/weinajin/multimodal\_explanation}}, for BraTS dataset, we trained a VGG-like 3D CNN with six convolutional layers. It receives multi-modal 3D MR images $X \in \mathbb{R}^{4 \times 240 \times 240 \times 155}$. We report evaluation results on the test set in a five-fold cross-validation.
We used a weighted sampler to handle the imbalanced data. The models were trained with a learning rate $= 0.0005$, batch size = 4, and training epoch of 32, 49, 55, 65, 30 for each fold selected by the validation data.
The accuracies of the five folds were $87.81 \pm 3.40\%$~(mean$\pm$std). 

For the synthetic brain tumor dataset, we fine-tuned a pre-trained DenseNet121 model that receives 2D mutli-modal MRI input slices of $X \in \mathbb{R}^{4 \times 256 \times 256}$.
We used the same training strategies as described above. The model achieved $95.70 \pm 0.06\%$ accuracy on the test set.

\section{Physician User Study}\label{user_study}
\begin{figure}[h]
    \centering
    \includegraphics[width=0.95\linewidth]{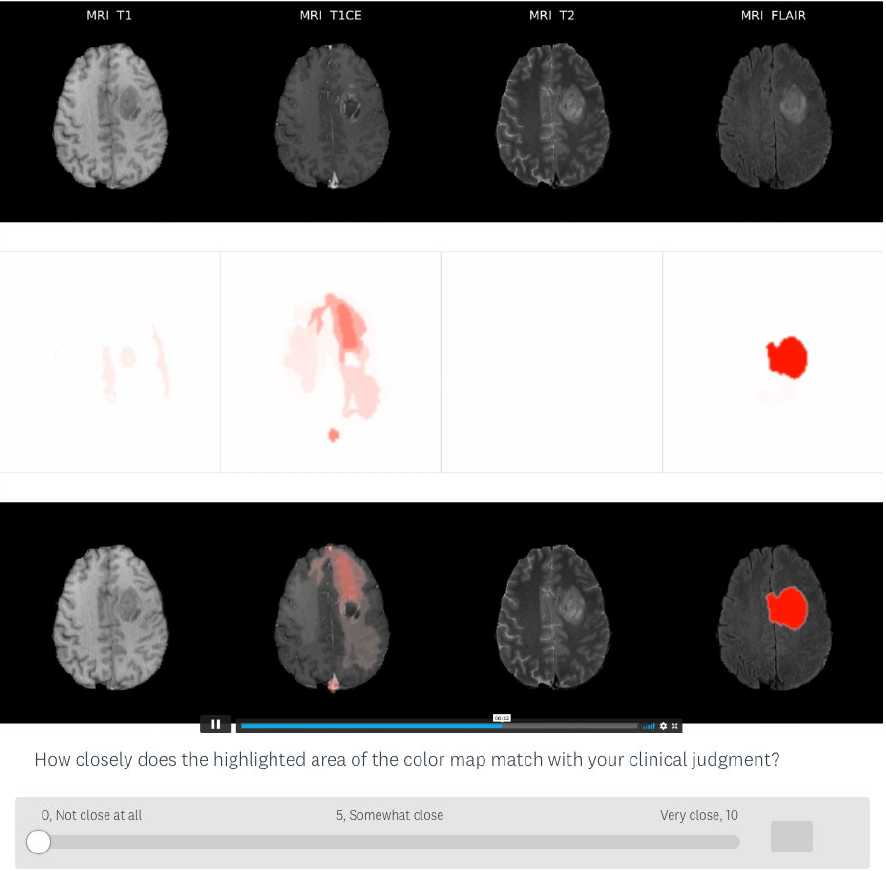}
    \caption{\textbf{3D heatmap (in video format) and questionnaire in the user study}. Column: each MRI modality. Row: MRI, heatmap, and heatmap overlaid on MRI. Redness indicates the importance of that area for prediction.}
    \label{fig:user_study}
\end{figure}

We conducted a user study including an online survey and an optional within-/post-survey interview with neurosurgeons. The survey, with a low-fidelity XAI system embedded to mimic XAI usage scenario in clinical decision support, asked neurosurgeons to interpret, comment, and rate the generated 3D multi-modal heatmaps (Fig.~\ref{fig:user_study}).
Neurosurgeons rated the heatmaps regarding \textit{explanation plausibility}, i.e.: ``how closely the highlighted areas of the heatmap match with your clinical judgment?'' The user study was approved by the Research Ethics Board of Simon Fraser University (Ethics No.: H20-03588).
Six neurosurgeons were recruited and participated in the survey.  
Two of them participated in the optional interview. The survey lasted for 1 hour, and the interview lasted for 30 minutes. Details on the user study are in the Appendix.

We extracted \textbf{clinical interpretation patterns of multi-modal explanations} using qualitative data analysis on clinicians' comments. When interpreting multi-modal images, physicians tend to \textbf{prioritize modalities} for a given task.

\begin{displayquote}
\textit{``Many of us just look at FLAIR and T1C. 
90\% of my time (interpreting the MRI) is on the T1C, and then I will spend 2\% on each of the other modalities.''} 
\end{displayquote}

In addition to describing the modality importance information, clinicians expected the heatmaps to correctly \textbf{localize features} that are discriminative for prediction (in this glioma grading task, it is the features inside the tumor regions~\cite{Law1989, Cho2018}). 

\begin{displayquote}
\textit{``This one (feature ablation heatmap, Fig.~\ref{fig:user_study}) is not bad on the FLAIR (modality), it (the tumor) is very well detected. I wouldn't give it a perfect mark, because I would like it to prioritize the T1C (modality) instead. But I'll give it (a score of) 75 (out of 100).''} 
\end{displayquote} 

We further propose computational evaluation metrics based on the clinical patterns (\S\ref{multimodal_eval}).

\section{Evaluation on Multi-Modal Explanations}\label{multimodal_eval}

\begin{figure*}[!ht]
    \centering
    \includegraphics[width=1\textwidth]{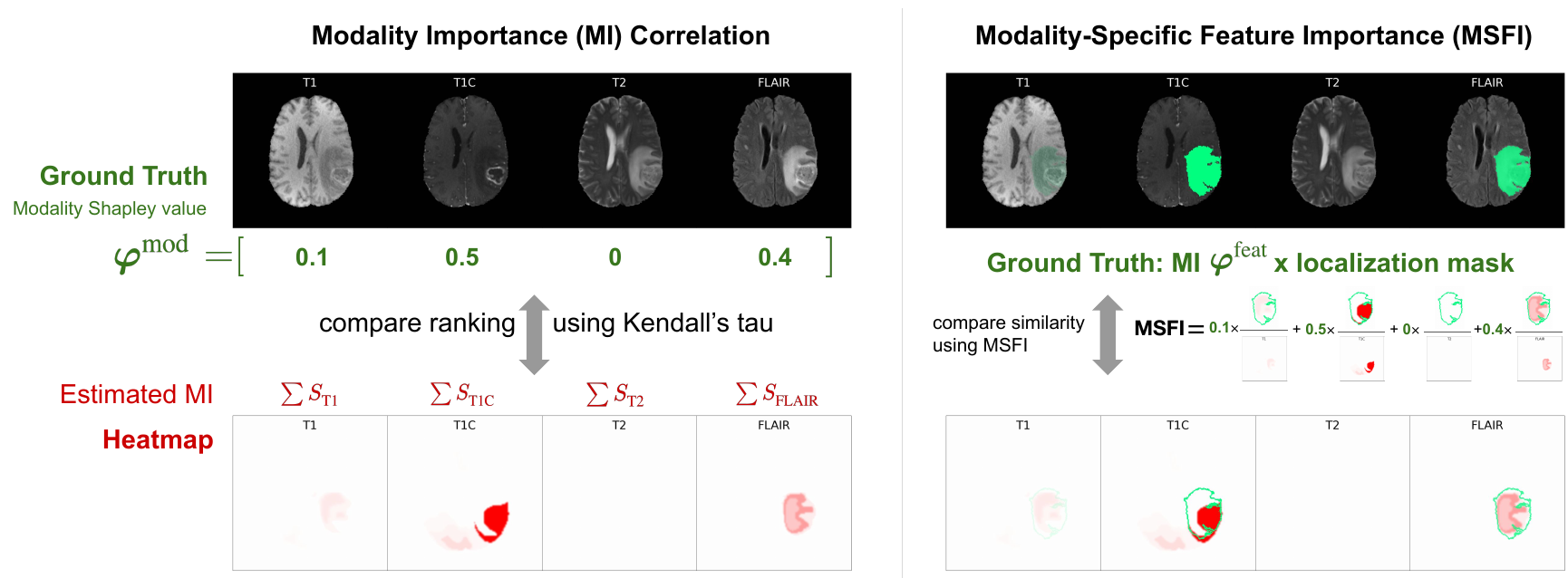}
    \caption{\textbf{Two proposed computational evaluation metrics for multi-modal medical image explanation.
    }}
    \label{fig:eval_outline}
\end{figure*}

Our evaluation focuses on \textbf{\textit{post-hoc}} XAI algorithms. Compared to \textit{ante-hoc} ones -- such as attention mechanism -- the evaluation results are not confounded by model types. Post-hoc XAI algorithms explain for already deployed or trained black-box models by probing model parameters and/or input-output pairs.
We include 16 post-hoc XAI algorithms in our evaluation, which belong to three categories:
\begin{itemize}
    \item \textbf{Activation-based}: GradCAM~\cite{8237336}
    \item \textbf{Gradient-based}: Gradient~\cite{simonyan2014deep}, Guided BackProp~\cite{springenberg2015striving}, Guided GradCAM~\cite{8237336}, DeepLift~\cite{10.5555/3305890.3306006}, InputXGradient~\cite{shrikumar2017just}, Integrated Gradients~\cite{10.5555/3305890.3306024}, Gradient Shap~\cite{NIPS2017_8a20a862},
Deconvolution~\cite{10.1007/978-3-319-10590-1_53}, Smooth Grad~\cite{smilkov2017smoothgrad}
    \item \textbf{Perturbation-based}: Occlusion~\cite{10.1007/978-3-319-10590-1_53,DBLP:conf/iclr/ZintgrafCAW17}, Feature Ablation, Shapley Value Sampling~\cite{CASTRO20091726}, Kernel Shap~\cite{NIPS2017_8a20a862}, Feature Permutation~\cite{JMLR:v20:18-760}, Lime~\cite{10.1145/2939672.2939778} 

\end{itemize}
A detailed review of these algorithms and heatmap post-processing method are in the Appendix. Next, we describe the metrics to evaluate multi-modal explanation. Corresponding to the above clinical patterns on multi-modal explanation, we propose two new evaluation metrics (Fig.~\ref{fig:eval_outline}) at different granularity levels: \textbf{1}) \textit{Modality importance (MI)}: it measures a model's overall importance of each modality as a whole; and \textbf{2}) 
\textit{Modality-specific feature importance (MSFI)}: it 
measures how well the heatmap can localize the modality-dependent important features in each modality.

\subsection{Modality Importance (MI)}\label{mod_shapley}
Corresponding to the clinical pattern on \textbf{modality prioritization}, MI uses importance scores to indicate how critical is a modality to the overall prediction. 
To determine the ground-truth MI, we use Shapley value from cooperative game theory~\cite{RM-670-PR}, due to its desirable properties such as efficiency, symmetry, linearity, and marginalism. In a set of $M$ modalities, Shapley value treats each modality $m$ as a player in a cooperative game play. It is the unique solution to fairly distribute the total contributions (in our case, the model performance) among each individual modality $m$. 

\paragraph{Shapley value-based MI ground-truth} We define the modality Shapley value $\varphi_m$ to be the ground truth MI value for a modality $m$.
It is calculated as: 

\begin{align}\label{eq1}
    \varphi_m(v)\!=\!\sum_{c \subseteq \mathcal{M} \backslash\{m\}} \!\frac{|c| !(M-|c|-1) !}{M !}(v(c \cup\{m\})-v(c)),
\end{align}
where $\mathcal{M} \backslash\{m\}$ denotes all modality subsets $\mathcal{M}$ not including modality $m$, and $v$ is the model performance metric. In our evaluation, we defined $v$ as the test set accuracy, and the accuracy on a subset of modalities $v(c)$ and $v(c \cup\{m\})$ was calculated by setting all values to $0$ for modalities that were not included in the subset. We denote such modality Shapley value as $\varphi_m^{\text{mod}}$. 

\paragraph{MI correlation} To measure the agreement on modality importance between heatmaps and the ground-truth modality Shapley value, for each post-processed heatmap, 
we calculate a vector of \textit{estimated MI} as the sum of all positive values of the heatmap for each modality. \textit{MI correlation} measures the MI ranking agreement between the ground-truth $\varphi^{\text{mod}}$ and the \textit{estimated MI}, calculated using Kendall's Tau-b correlation.
MI correlation is a measure of \textit{explanation faithfulness}, since the ground-truth Shapley value reflects the model's internal decision process, and is calculated without human prior knowledge.

\subsection{Modality-Specific Feature Importance (MSFI)}

MI prioritizes the important modality, but it is a coarse measurement and does not examine the particular image features within each modality. 
We further propose MSFI metric that corresponds to the clinical pattern on both \textbf{feature localization} and \textbf{modality prioritization}. MSFI combines two ground-truth information: MI and modality-dependent important features. Specifically, for each modality $m$, MSFI calculates the portion of heatmap values $S_m$ inside the ground truth feature localization mask $L_m$, weighted by MI  $\varphi_m$ (which is normalized to $[0, 1]$):

\begin{align*}
\widehat{\text{MSFI}} &=\sum_{m}  \varphi_m \frac{ \sum_i \mathbbm{1}
( L_m^i >0 ) \odot S_m^i }{ \sum_i 
S_m^i  },\\
\text{MSFI} &= \frac{\widehat{\text{MSFI}}} {\sum_{m} \varphi_m},
\end{align*}
where 
$i$ denotes the spatial location of heatmap $S_m$;
$\mathbbm{1}$ is the indicator function that selects heatmap values inside feature mask $L_m$; 
$\widehat{\text{MSFI}}$ is unnormalized, and ${\text{MSFI}}$ is the normalized metric in the range $[0, 1]$. 
A higher MSFI score indicates a heatmap is better at highlighting important modalities and their localized features. 
If the feature localization annotations $L_m$ reflects human prior knowledge on the given task, then MSFI is a metric on explanation \textit{plausibility}. Otherwise if $L_m$ reflects the intrinsic knowledge the model learned (as in the later subsection on a synthetic dataset), MSFI can also be used as a metric for \textit{faithfulness}. Unlike other ground-truth similarity metrics such as IoU, MSFI is less dependent on either heatmap signal intensity, or area of the ground truth localization mask, which makes it a robust metric. Next, we describe our evaluation experiments of applying MSFI on a real dataset (BraTS) and a synthetic dataset.%

\subsubsection{MSFI Evaluation on a Medical Image Dataset of Real Patients}
We use the same BraTS data and model as in \S\ref{mod_shapley}.
To make the ground truth represent the modality-specific feature localization information, we slightly change the way to compute the modality Shapley value $\varphi_m$. We define $\varphi_m^{\text{feat}}$ to be the importance of the localized feature on each modality. Specifically, instead of ablating the modality as a whole to create a modality subset $c$ in Eq.~\ref{eq1}, we zero-ablate only the localized feature region defined by the feature localization map. We calculate MSFI score with the new ground truth $\varphi_m^{\text{feat}}$.

\subsubsection{MSFI Evaluation on a Synthesized Dataset with Controllable Ground Truths}\label{syn}

To better control the ground truths of modality importance and feature localization, we use a synthetic multi-modal medical image dataset on the same brain tumor grading task. 
To control the ground-truth of feature localization, we use the GAN-based (generative adversarial network) tumor synthesis model developed by~\citet{Kim2021} to generate two types of tumors and their segmentation maps, mimicking low- and high-grade gliomas  by varying their shapes (round vs. irregular~\cite{Cho2018}). 

To control the ground-truth of modality importance, inspired by~\citet{pmlr-v80-kim18d}, 
we set tumor features on T1C modality to have 100\% alignment with the ground-truth label, and on FLAIR to have a probability of 70\% alignment, i.e., the tumor features on FLAIR corresponds to the correct label with 70\% probability. The rest modalities have 0 MI value, as they are designed to not contain class discriminative features. The model may learn to pay attention to either the less noisy T1C modality, or the more noisy FLAIR modality, or both. To determine their relative importance as the ground truth MI, 
we test the well-trained model on two datasets that only show tumors (without the background of normal brain tissue) in all modalities:

    $\bullet$ \textit{TIC dataset}: The tumor shape feature has \textit{100\%} alignment with the ground-truth label in \textit{T1C} modality, and 0\% alignment in FLAIR. Its test accuracy is denoted as $\text{Acc}_{\text{T1C}}$.
    
    $\bullet$ \textit{FLAIR dataset}: The tumor shape feature has \textit{100\%} alignment with ground truth in \textit{FLAIR} modality, and 0\% alignment in T1C. Its test accuracy is denoted as $\text{Acc}_{\text{FLAIR}}$.

The test performance $\text{Acc}_{\text{T1C}}$ and $\text{Acc}_{\text{FLAIR}}$ indicate the degree of model reliance on that modality to make predictions. We use them as the ground truth MI. 
On the test set, $\text{Acc}_{\text{T1C}}=0.99$, $\text{Acc}_{\text{FLAIR }}=0$. In this way, we constructed a model with known ground truth of MI = 1 for T1C, and 0 for the rest modalities. The MSFI in this case is a metric for \textit{faithfulness}, as both ground truths are known and baked in the model decision process. We then generate heatmaps on top of the model, and calculate their MSFI.

\begin{table*}[]
\small
    \centering
    \resizebox{\textwidth}{!}{
    \begin{tabular}
    {cccccccccc}
    \toprule
     & \textcolor{orange}{\textbf{\dotuline{MSFI} (\dotuline{BraTS})}} & \textcolor{orange}{\textbf{\dotuline{Stat. Sig.}}} & \textcolor{blue}{\textbf{\underline{MSFI} (\underline{Synthetic})}}&
    \textcolor{blue}{\textbf{\underline{MI}} \underline{\textbf{Correlation}}}
     & \textcolor{blue}{\underline{\textbf{diffAUC}}} & \dotuline{\textcolor{orange}{\textbf{FP}}} & \dotuline{\textcolor{orange}{\textbf{IoU}}} & \textcolor{orange}{\textbf{\dotuline{Doctors'} \dotuline{Rating}}} & \textbf{Speed} (second)\\
        \hline
Guided BackProp
& 0.48$\pm$0.33& NS & \textbf{0.49$\pm$0.21} & \textbf{0.80$\pm$0.27} & 
0.21$\pm$0.24 & 
0.34$\pm$0.29 & 
0.02$\pm$0.01 &
0.6$\pm$0.1  & \textbf{1.7$\pm$1.1}\\
\hline
Guided GradCAM
& 0.50$\pm$0.36 & $ \star\star$  & \textbf{0.42$\pm$0.29} & \textbf{0.81$\pm$0.26} & 
0.26 $\pm$ 0.25 & 
0.37$\pm$0.31 & 
0.02$\pm$0.02 &
0.1$\pm$0.0 & 2.2$\pm$1.4\\
\hline

InputXGradient
& 0.51$\pm$0.32 & $ \star$  & 0.23$\pm$0.14 & \textbf{0.87$\pm$0.16} & 0.17 $\pm$ 0.12 &
0.40$\pm$0.30 & 
0.08$\pm$0.05 &
0.1$\pm$0.0  & \textbf{1.7$\pm$1.1}\\
\hline
DeepLift
& 0.54$\pm$0.34 & $ \star$  & 0.22$\pm$0.23 & 0.53$\pm$0.45 & 0.19 $\pm$ 0.14 
& 0.43$\pm$0.32 & 
0.08$\pm$0.05 & 0.6$\pm$0.2  &
3.8$\pm$2.0\\
\hline

Integrated Gradients
& 0.48$\pm$0.31& $ \star$  & 0.22$\pm$0.19 & 0.73$\pm$0.39 & 0.17 $\pm$ 0.12 & 
0.36$\pm$0.28 & 
0.08$\pm$0.05 &
0.5$\pm$0.0  & 
62$\pm$29\\
\hline
Occlusion
& 0.28$\pm$0.26& $ \star\star\star$  & 0.22$\pm$0.25 & 0.60$\pm$0.33 & 0.13 $\pm$ 0.15 & 
0.18$\pm$0.19 & 
0.03$\pm$0.02 &
0.6$\pm$0.2  & 989$\pm$835\\
\hline
Gradient Shap
& 0.48$\pm$0.31& $ \star$  & 0.22$\pm$0.19 & 0.53$\pm$0.40 & 0.17 $\pm$ 0.12 & 
0.36$\pm$0.28 & 
0.08$\pm$0.05 &
0.5$\pm$0.0  & 6.8$\pm$3.0 \\
\hline
Feature Ablation& 0.48$\pm$0.30& $ \star\star\star$  & 0.19$\pm$0.23 & 0.27$\pm$0.44 & \textbf{0.30 $\pm$ 0.15} & 
0.35$\pm$0.28 & 
0.05$\pm$0.06 &
0.4$\pm$0.4  & 74$\pm$23\\
\hline
Gradient
& 0.34$\pm$0.23& NS & 0.19$\pm$0.13 & 0.47$\pm$0.16 & 0.05 $\pm$ 0.09 &
0.20$\pm$0.16 & 
0.02$\pm$0.01 &
0.6$\pm$0.6  & 1.8$\pm$1.1\\ 

\hline
Shapley Value Sampling
& 0.38$\pm$0.24& $ \star\star\star$  & 0.10$\pm$0.10 & 0.47$\pm$0.65 & \textbf{0.35 $\pm$ 0.04} & 
0.25$\pm$0.21 & 
0.04$\pm$0.05 &
0.2$\pm$0.1  & 2018$\pm$654\\
\hline
Kernel Shap
& 0.28$\pm$0.25& $ \star\star$  & 0.08$\pm$0.08 & NaN & 0.26 $\pm$ 0.16 & 
0.18$\pm$0.20 & 
0.06$\pm$0.08 &
0.1$\pm$0.0  & 194$\pm$100\\
\hline
Feature Permutation
& 0.23$\pm$0.26& NS & 0.08$\pm$0.07 & NaN & 0.05 $\pm$ 0.05 & 
0.13$\pm$0.18 & 
0.05$\pm$0.07 &
0.1$\pm$0.0  & 14$\pm$2.2\\
\hline
Lime
& 0.24$\pm$0.21& $ \star\star$   & 0.05$\pm$0.07 & 0.53$\pm$0.58 & \textbf{0.37 $\pm$ 0.08} & 
0.14$\pm$0.16 & 
0.05$\pm$0.06 &
0.1$\pm$0.0 & 341$\pm$181\\
\hline
Deconvolution
& 0.26$\pm$0.23& NS & 0.04$\pm$0.02 & 0.73$\pm$0.39 & 0.11 $\pm$ 0.21 & 0.17$\pm$0.17 & 
0.02$\pm$0.01 & 0.4$\pm$0.4 & 
1.8$\pm$1.0 \\
\hline
Smooth Grad
& 0.27$\pm$0.17& $ \star$  & 0.03$\pm$0.02 & 0.67$\pm$0.00 & 0.29 $\pm$ 0.25 & 0.16$\pm$0.12 & 
0.02$\pm$0.01 &
0.7$\pm$0.1 & 12$\pm$6\\

\hline
GradCAM
& 0.04$\pm$0.03& $\star\star\star$ & 0.02$\pm$0.02 & NaN & 
0.16 $\pm$ 0.19 & 0.02$\pm$0.01 & 
0.02$\pm$0.01 &
0.0$\pm$0.0  & \textbf{0.6$\pm$0.3}\\
\bottomrule
    \end{tabular}}
    \caption{\textbf{The evaluation results}.
    The table shows mean $\pm$ std for each XAI algorithm regarding different evaluation metrics on the test set. Metrics are in the range $[0, 1]$ (except for diffAUC and MI which is $[-1, 1]$), the higher, the better. Metrics for \underline{\textcolor{blue}{faithfulness}} and \dotuline{\textcolor{orange}{plausibility}} are marked respectively, with \textbf{bolded} text indicating the top faithfulness performance for a metric. Stat. Sig. tests the correlation between MSFI (BraTS) score and the two groups of correct/incorrect predictions, with $\star$ indicates $p<0.05$; $\star\star$ for $p<0.01$; and $\star\star\star$ for
    $p<0.001$; NS for not significant. ``NaN'' in MI is because the heatmap is not modality-specific and the correlation is not computable. Speed is the time spent to generate a heatmap.
    }
    \label{tab:eval_result}
\end{table*}

\section{Evaluation Results}

\subsection{Modality Importance Correlation}
The MI correlation results are shown in Table~\ref{tab:eval_result}.
Except for certain XAI algorithms (GradCAM, KernalSHAP, Feature Permutation) that could only generate one same heatmap explanation for all modalities (non-modality-specific heatmap), in general, most algorithms correctly identified the important modalities for model decision, indicating they were \textit{faithful} at the modality level, but with large variances among individual data points.

\subsection{Modality-Specific Feature Importance}

\subsubsection{MSFI Results on BraTS Dataset}

For all heatmap algorithms, their mean MSFI scores are in the middle to lower range, with large variances among individual data points (Table.~\ref{tab:eval_result}, Fig.~\ref{fig:metric}-top). To test whether human judgment of heatmap \textit{plausibility} measured by MSFI can be indicative of the model's decision quality, we divided the data into correctly or incorrectly predicted groups. For each algorithm, we tested whether there was a significant difference regarding their MSFI scores between the two groups using Mann-Whitney U test, and the significant level for each algorithm is shown in Table~\ref{tab:eval_result}. Despite some algorithms showed statistical significance, such significance may not be easily captured by clinical users, 
since the visualized MSFI distributions of the correctly and incorrectly predicted groups were overlapping with each other (as shown by the blue and red dots in Fig.~\ref{fig:metric}-top). This indicates assessing heatmap quality may not be a reliable signal for physicians to get alerted to AI model's potential decision flaw. This finding echoes with the prior study on the heatmaps' poor diagnosing capability for model generalization~\cite{viviano2021saliency}.  In summary, all examined algorithms did not fulfill the clinical requirement of being indicative of model decision quality via users' \textit{plausibility} judgment.

\subsubsection{MSFI Results on the Synthetic Dataset} 
\begin{figure*}[!tb]
    \centering
    \includegraphics[width=0.95\textwidth]{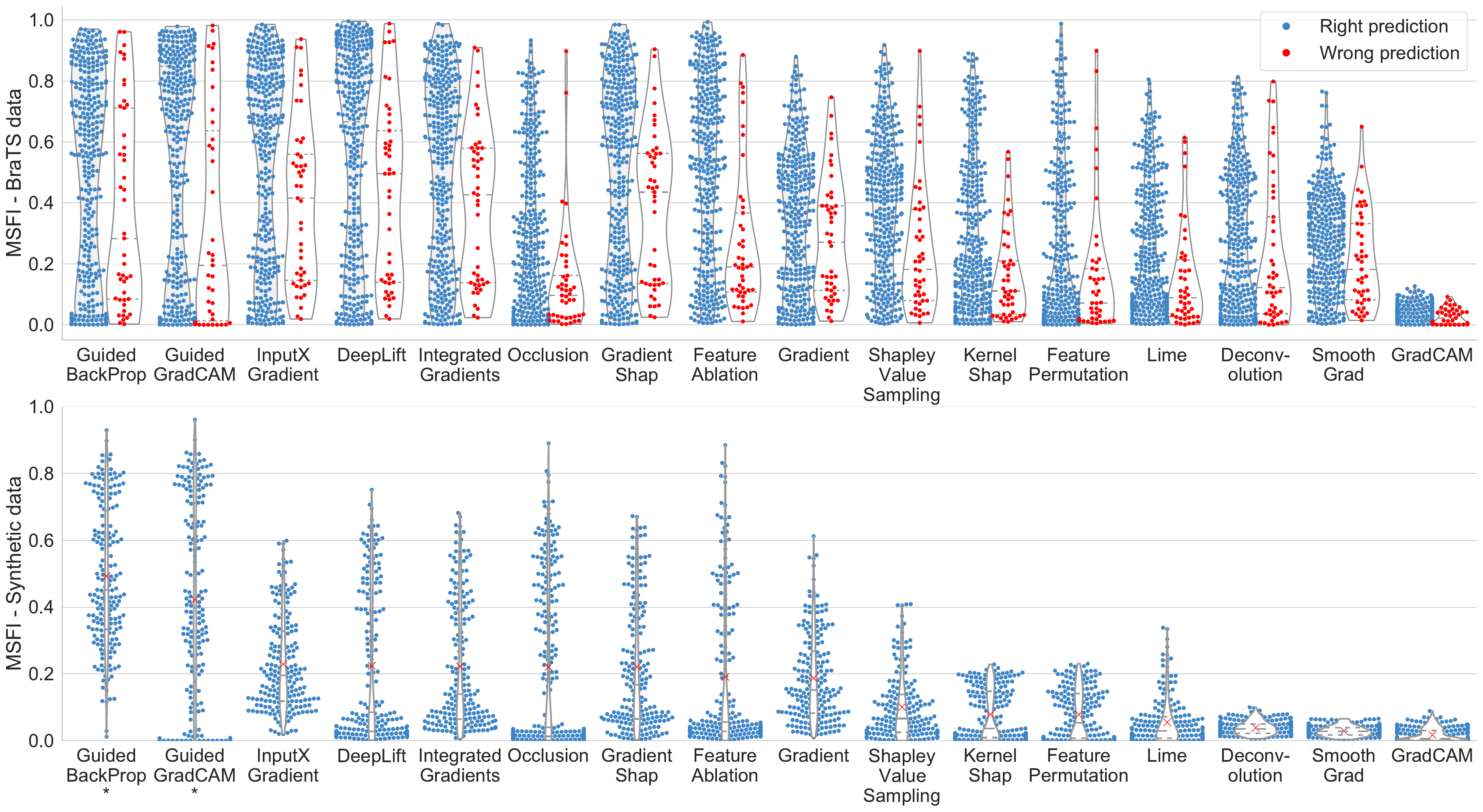}
    \caption{\textbf{MSFI scores of the evaluated 16 heatmap algorithms.} 
    The swarm and its overlaid violin plots show the evaluation score distribution for MSFI on BraTS dataset (top), and MSFI on synthetic dataset (bottom). X-axis is each heatmap method. Y-axis is the MSFI score, with a higher score means better alignment of a heatmap with the clinical patterns on modality prioritization and feature localization. 
    For the bottom plot, red cross indicate the mean. There was a statistically significant difference among the 16 heatmap methods for each subplot as determined by Friedman test, and the top methods have their name marked with $*$ (determined by not significantly different from the top two means using post-hoc Nemenyi test).}
    \label{fig:metric}
\end{figure*}

With known ground truths on important modalities and features to assess \textit{faithfulness} at the fine-grained feature level, the results (Table~\ref{tab:eval_result}, Fig.~\ref{fig:metric}-bottom) showed the examined algorithms had poor agreements with the ground truths, indicating their low \textit{faithfulness} performances at the modality-specific feature level: even for algorithms that had relatively higher MSFI, the scores were less than 0.5. In addition, their faithfulness performances were not stable, with large variances across data points.

\subsection{Comparison with Non-Modality-Specific Evaluations}\label{other_metric}
We utilized additional existing metrics that are non-modality specific.
To evaluate \textit{faithfulness}, we iteratively ablated the input from the most to the least important features according to the heatmap, and plotted the relationship between gradual feature ablation and model accuracy. We used \textbf{$\text{diffAUC}$} to quantify the degree of performance deterioration by calculating the difference of area under the curve (AUC) between an XAI algorithm and its random ablation baseline. The low diffAUC scores in Table~\ref{tab:eval_result} indicates a low level of \textit{faithfulness} at the non-modality-specific feature level.

To quantify non-modality-specific \textit{plausibility}, we used metrics of \textbf{IoU} and feature portion (\textbf{FP}, the sum of heatmap values inside the ground-truth feature mask over the total values).
We inspected the relationship between MSFI and the non-modality-specific metrics: MSFI had a strong Pearson correlation (0.96) with FP, and a moderate correlation (0.55) with IoU. MSFI can be regarded as a generalized form of FP: it requires the same ground-truth annotation as of IoU and FP,
and has the advantage to incorporate the clinical interpretation pattern of modality prioritization.  

Physicians' average quantitative rating on heatmap quality had moderate correlation with MSFI (0.43) and FP (0.43), and low correlation with IoU (0.26).
In addition, physicians' inter-rater agreement
on the heatmap quality is low (Krippendorff's Alpha = 0.16, Fleiss' kappa = -0.017),
indicating that doctors' judgment of heatmap quality could be very subjective.

\section{Discussions}
\subsubsection{Existing XAI Algorithms Failed to Fulfill Clinical Requirements} 
The examined 16 post-hoc heatmap algorithms failed to meet the clinical requirements, as they were not \textit{faithful} to the model decision process at the feature level (evidenced by the low and unstable MSFI on synthetic data and diffAUC), and users' \textit{plausibility} judgment of the explanation was not indicative of the AI model decision quality (evidenced by the MSFI (BraTS) statistical test and its distribution visualization).  
The poor and instable explanation performance may lead to undesirable consequences in clinical settings.
For example, in our user study, we observed doctors tend to assume the explanation is totally \textit{faithful} to the model's decision process (which aligns with prior findings~\cite{10.1145/3313831.3376219}), therefore would take or reject the model's suggestion by judging the \textit{plausibility} of the explanation. 
Given doctors' mental model on the assumption of totally \textit{faithful} explanation, the evaluation for \textit{faithfulness} should be put ahead of the evaluation for \textit{plausibility} on model decision quality indication.
Future work needs to propose explanation methods that are both \textit{faithful} to the model decisions, and the \textit{plausibility} assessment is more indicative of model decision quality.

\subsubsection{Real-World Application of MSFI}
The proposed MSFI can be regarded as a proxy for physicians' manual assessment of heatmap quality.
By automatically quantifying human assessment, MSFI helps AI practitioners to automate the XAI algorithm selection/optimization process to identify algorithms whose plausibility measure has a high correlation with model decision quality.
MSFI can be applied to other multi-modal medical image tasks. It requires feature masks on a batch of test data which can be generated by human annotation or trained models. Depending on the task, the feature masks could be modality-specific or the same for all modalities (as in the case of BraTS dataset). 
The MSFI metric is the first step towards tackling the clinically important problem of multi-modal explanation.

\section{Conclusion}
Explainable AI is an indispensable component when implementing AI as a clinical assistant on medical image-related tasks. In this work, we investigate the essential question raised in both machine learning and clinical fields: Can existing XAI methods fulfill clinical requirements on multi-modal explanation? We conduct a clinical requirement-grounded, systematic evaluation to answer this question, including both computational and physicians' assessment. By incorporating physicians' clinical image and explanation interpretation patterns into the evaluation metric, we propose MSFI that encodes both modality prioritization and feature localization.
Our evaluation with 16 existing XAI algorithms on a brain tumor grading task shows that, the existing XAI algorithms are not proposed to fulfill the clinical requirements on multi-modal medical imaging explanation.
This work sheds light on the risk of directly applying post-hoc XAI methods on black-box models in multi-modal medical imaging tasks. Future work may incorporate the informativeness of MSFI for decision quality into the objective function, and propose new heatmap methods to truthfully reflect the model decision process and quality.
Ultimately, we expect this study can help increase awareness of trustworthiness in AI for clinical tasks and accelerate other applications of AI for healthcare.

\section*{Acknowledgments}
We thank all physician participants in the user study. We thank Mostafa Fatehi, Sunho Kim, Yiqi Yan, Mayur Mallya, Shahab Aslani, and Ben Cardoen for their generous support and helpful discussions. We thank all reviewers for their comments. This study was funded by BC Cancer Foundation-BrainCare BC Fund, and was enabled in part by support from NVIDIA and Compute Canada (www.computecanada.ca).

\begin{appendices}

\section{User Study Details}
The user study consists of an online survey and an optional within-/post-survey interview. The heatmap reading survey was to try to mimic real-world usage scenarios in clinical decision-support settings. In the survey, after informed consent, the physician participants were first introduced to the AI model and got to know its performance on the test set. The participants then used the AI model on a new patient MRI, and inspected AI's prediction and its explanation. 
The MRI and its heatmaps were 3D images presented in axial view and slice-by-slice in video format. We used the test data, the trained model of Fold 1, and the model’s prediction as the target label to generate the heatmaps. We randomly selected one MRI and their heatmaps for each correctly or incorrectly predicted case.

The interview session was one-on-one, remote, open-ended, semi-structured. The interview was conducted during or after the survey session, depending on the participant's preference. 
We video-recorded the interview session, and analyzed the transcribed qualitative interview data using thematic analysis~\cite{Braun2012}. 

We recruited participants by directly contacting the researchers' clinical collaborators and snowball sampling in a local general hospital. The inclusion criteria were: the participant must hold a Doctor of Medicine degree, and work in neurosurgery, radiology, or neuro-radiology specialty. The participants were thanked with \$100 CAD for their time and effort in the study. The recruited participants were five neurosurgical residents and one neurosurgical fellow, with an average age of $30.2 \pm 1.5$ (mean ± std), female : male $=1:4$, years of practicing medicine: $4.6 \pm 3.2$, and years of practicing neurosurgery specialty: $3.2 \pm 2.6$. For their level of familiarity with AI, 3 participants (50\%) can code but not write AI code; 1 participant (17\%) uses AI in work or life (such as using face recognition to unlock phone, and Siri), and 2 participants (33\%) only hear of AI. For their attitude toward incorporating AI in medical practice, 4 participants (80\%) hold a positive attitude, while 1 participant (20\%) is skeptical. Participants also rated the potential usage scenarios for AI explanations, as listed in Table~\ref{tab:utility}.

\begin{table*}[!h]
    \centering
    \begin{tabular}{l|p{6em} p{10em}}
    \toprule
Motivations to Check Explanations                 &   Selected by \% Participants & Individual Participant’s Rankings\\
\hline
\textbf{When I doubt about the prediction from AI}       & 83\% & 1, 2, 2, 3, 4\\
\textbf{To verify AI's decisions}                        & 83\%                & 2, 2, 3, 3, 5\\
\textbf{To build and calibrate my trust in this AI }   & 50\%        & 1, 1, 3\\
\textbf{To ensure the safety use of AI }           & 50\%           & 1, 5, 6\\
\textbf{For a difficult case, when I am not certain} & 50\%         & 2, 5, 5\\\hline
To learn from AI                                & 33\%                     & 4, 4\\
To make differential diagnosis               & 16\%                 & 1\\
To make new medical discovery                    & 16\%           & 3\\
When I am trading off among multiple objectives for my patient   & 16\%  & 4\\
Before discussion with my colleagues                    & 16\%                         & 6\\\hline
To meet the ethical requirements                 & 0\%                    & \\
To meet the legal requirements                          & 0\%                    & \\
To ensure fairness and no biases in the AI model        & 0\%                    & \\
To generate report or patient chart                        & 0\%                    & \\
To improve my patients' outcomes                       & 0\%                    & \\
\bottomrule
    \end{tabular}
    \caption{\textbf{Clinical Utilities of XAI}. Participants selected and ranked the potential motivations to check AI's explanations for the survey question "When are you most likely to check those color map (heatmap) explanations from AI?". The utilities are ranked by the frequency of being selected. The most selected utilities are bolded. The numbers in the 3rd column represent each participant's ranking of a motivation when it was selected by a participant.}
    \label{tab:utility}
\end{table*}

\section{Review of XAI Algorithms}

We include 16 commonly-used gradient- and perturbation-based heatmap methods. We give a brief review for each of them.

\subsection{Gradient- and Activation-Based Heatmap Methods}
\subsubsection*{CAM} CAM (Class Activation Mapping)~\cite{7780688} generates a heatmap for a prediction by aggregating the internal activations of a neural network layer and weighting each neuron's weights in that layer to the final decision layer.

\subsubsection*{Grad-CAM} It is similar to CAM but replaces the weights with gradients of the target prediction with respect to the activation map~\cite{8237336}. We only include Grad-CAM in our evaluation, because it does not require special model architecture as CAM does.

For activation-based methods including CAM and Grad-CAM, because the activation maps at a deeper layer could not reflect the modality-specific information, which is aggregated at the first convolutional layer, the output heatmap is a single-modality image, which is not modality-specific. We copy such a heatmap to all modalities to compare it with other methods.

\subsubsection*{Gradient} It reflects how quickly the output changes when input changes~\cite{simonyan2014deep}.

\subsubsection*{Input $\times$ Gradient} It multiplies the input by the gradient signal to obtain a first-order Taylor approximation~\cite{shrikumar2017just}.
Compared with Gradient, it tends to produce sharper heatmaps. Since \cite{shrikumar2017just} showed layer-wise relevance propagation (LRP) is equivalence to Input $\times$ Gradient when all activations are piece-wise linear and biases are included, we only include Input $\times$ Gradient in the evaluation.

\subsubsection*{SmoothGrad} It smooths the noisy gradient signals by averaging the heatmaps for an input and its random neighborhood samples~\cite{smilkov2017smoothgrad}.

\subsubsection*{Deconvolution} It modifies the gradient computation rule at ReLU activation function. Instead of back-propagating non-negative \textit{input} gradients as in the vanilla Gradient method, Deconvolution only back-propagates non-negative \textit{output} gradients~\cite{10.1007/978-3-319-10590-1_53}. 

\subsubsection*{Guided Backpropagation} It combines Gradient and Deconvolution methods by back-propagating \textit{input} and \textit{output} gradients that are both non-negative~\cite{springenberg2015striving}. 

\subsubsection*{Guided Grad-CAM} It  computes the element-wise product between Guided Backpropagation and the up-sampled \& broadcasted Grad-CAM signal~\cite{8237336}.

\subsubsection*{Integrated Gradient} It approximates the path integral of gradients along the straight line from a neutral baseline input to the target input~\cite{10.5555/3305890.3306024}.

\subsubsection*{DeepLIFT} It explains the prediction difference from an uninformative baseline by introducing difference-from-reference operation in the backpropagation rule~\cite{10.5555/3305890.3306006}. DeepLIFT and Integrated Gradients both address the gradient ``saturation'' problem, and DeepLIFT is faster than Integrated Gradient as it only needs one backward pass to compute the explanation.

\subsubsection*{Gradient SHAP} It approximates Shapley values by computing the expectation of gradients~\cite{NIPS2017_8a20a862}.

\subsection{Perturbation-Based Heatmap Methods}

\subsubsection*{Occlusion} It occludes part of the image with a sliding window, and averages the output differences as the feature attribution~\cite{10.1007/978-3-319-10590-1_53,DBLP:conf/iclr/ZintgrafCAW17}. In our implementation, the occlusion is done modality-wise to generate heatmaps that are modality specific. The occluded regions are replaced by values drawn from a normal distribution with the same mean and standard deviation of the given input modality. We experimented with different sizes of sliding window and stride to balance heatmap resolution and computational time.

\subsubsection*{Feature Ablation} It is similar to Occlusion, but occludes the individual image features rather than using a sliding window. In our implementation, we use modality-wise superpixel segmentation masks as the image features to generate modality-specific heatmaps, and replace the ablated feature with baseline value of $0$s.

\subsubsection*{Feature Permutation}It replaces image feature by shuffling the feature values within a batch, and computes the prediction difference accordingly~\cite{JMLR:v20:18-760}. 

\subsubsection*{LIME} LIME (local interpretable model-agnostic explanation) learns an interpretable model by perturbing and sampling the neighbor data points around the input~\cite{10.1145/2939672.2939778}. 

\subsubsection*{Shapley Value Sampling} It relies on the concept of a feature's Shapley value, which is the average marginal feature attribution across all possible feature combination subsets~\cite{RM-670-PR}. Shapley Value Sampling is an efficient sampling method to overcome the expensive enumeration of all possible feature combinations~\cite{CASTRO20091726}.

\subsubsection*{Kernel SHAP} It uses the LIME framework to compute Shapley values~\cite{NIPS2017_8a20a862}. Since it only receives one superpixel feature segmentation mask that is shared across modalities, the produced heatmaps are \textit{not} modality-specific.

\section{Post-Processing Heatmaps}
Before evaluating and visualizing the heatmaps, we post-processed them by first capping the top 1\% outlying values (following~\cite{smilkov2017smoothgrad}). We focused on the positive values of the heatmaps as they are interpreted as evidence towards the model decision (a.k.a. importance scores). Since the negative values in the heatmaps may carry different meanings for different XAI algorithms, and since it is difficult for end-users to understand and interpret the negative values from our pilot study, in the evaluation pipeline, except for the visualization, we ignored the negative values by setting them to $0$.
We then scaled the values of the heatmaps to $[0,1]$, and applied a Gaussian kernel to visually smoothen the heatmaps. 
\end{appendices}
\bibliography{xai_eval}

\end{document}